\pdfoutput=1

\documentclass[11pt]{article}

\usepackage[preprint]{acl}

\usepackage{times}
\usepackage{latexsym}

\usepackage[T1]{fontenc}

\usepackage[utf8]{inputenc}

\usepackage{microtype}

\usepackage{inconsolata}

\usepackage{graphicx}

\usepackage{booktabs}
\usepackage{amsmath,amssymb}

\usepackage{algorithm}
\usepackage[noend]{algpseudocode}
\algrenewcommand\algorithmicdo{}

\usepackage[whole]{bxcjkjatype}
\usepackage{hyperref}

\usepackage{subcaption}

\title{Media of Langue:\\The Interface for Exploring Word Translation Network/Space}

\author{
  Goki Muramoto, Atsuki Sato, Takayoshi Koyama \\
  The University of Tokyo \\
  \texttt{goki.muramoto@gmail.com}
}

\begin{document}
\maketitle
\begin{abstract}
In the human activity of word translation, two languages face each other, mutually searching their own language system for the semantic place of words in the other language. We discover the huge network formed by the chain of these mutual translations as \textit{Word Translation Network}, a network where words are nodes, and translation volume is represented as edges, and propose \textit{Media of Langue}, a novel interface for exploring this network. \textit{Media of Langue} points to the semantic configurations of many words in multiple languages at once, containing the information of existing dictionaries such as bilingual and synonym dictionaries. We have also implemented and published this interface as a web application, focusing on seven language pairs. This paper first defines the \textit{Word Translation Network} and describes how to actually construct the network from bilingual corpora, followed by an analysis of the properties of the network. Next, we explain how to design a \textit{Media of Langue} using the \textit{Word Translation Network}, and finally, we analyze the features of the \textit{Media of Langue} as a dictionary. 
Our website is \url{www.media-of-langue.org}.


\end{abstract}

\section{Introduction}

\begin{figure*}[t]
\centering
\includegraphics[width=1\textwidth]{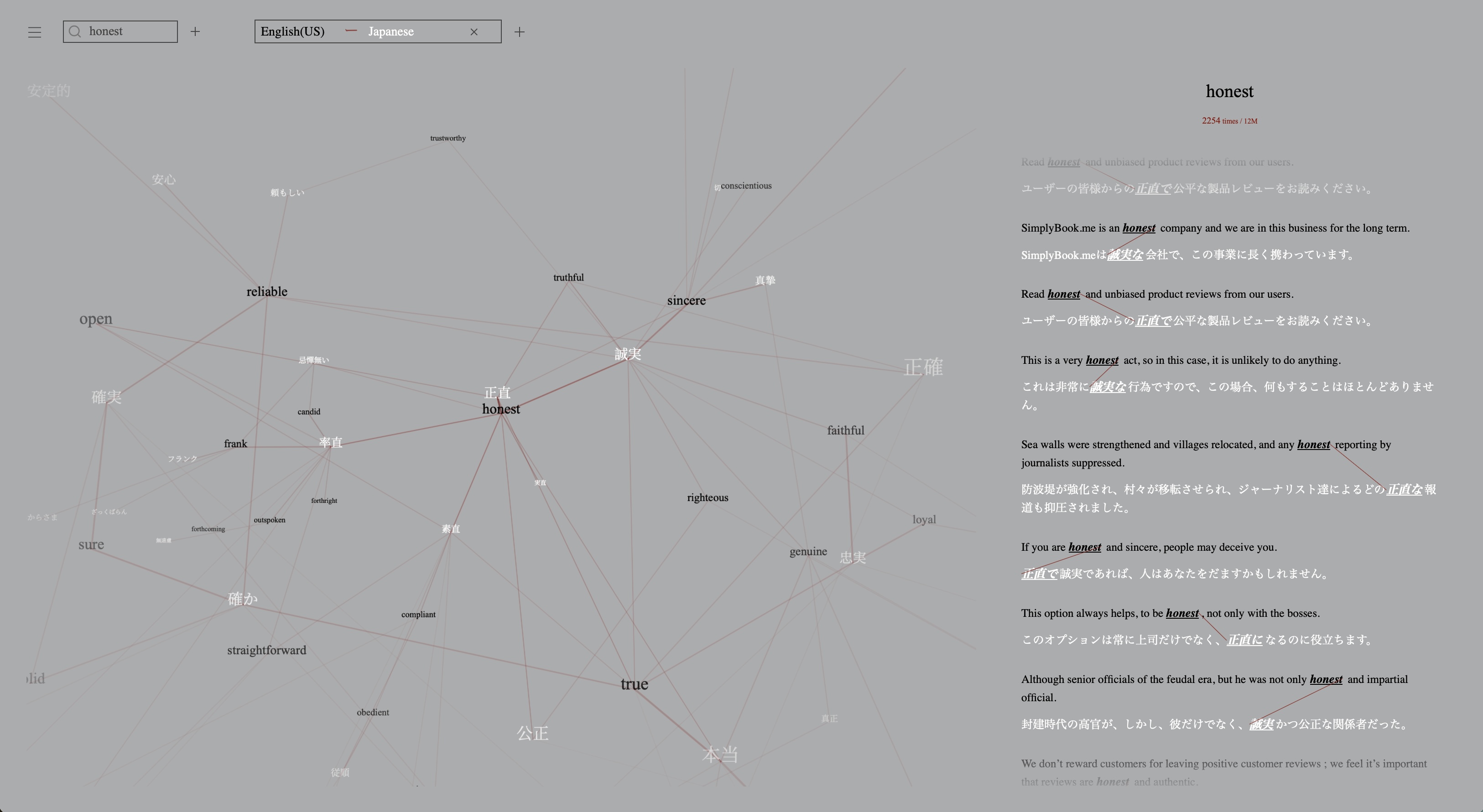}

\caption{A screenshot of the our implementation of \textit{Media of Langue}}
\label{fig:mediaOfLangue}
\end{figure*}



\begin{quote}

This word was translated into that word.

That word was translated into that one.
    
That one was...

What a vast network could be formed by collecting and connecting the countless word translations that have occurred in the past.
\end{quote}


Language exhibits both universality and diversity. We have faced this seemingly ordinary fact many times through the problem that “translation is neither completely possible nor completely impossible. Researchers have theoretically examined the balance between the possibility and impossibility of translation. The rejection of the notion that concepts precede names~\cite{jowett1892charmides, de1989cours} and the acknowledgment of linguistic relativity---in which language itself shapes perception~\cite{von1999humboldt, boas1938handbook, sapir2004language, whorf2012language}---have rendered the impossibility of translation a scientific issue that extends beyond everyday intuition. In particular, Quine raised the question of the ``indeterminacy of translation,'' believing that translation is necessarily polysemic and that it is impossible to transfer the meaning of one language exactly to another~\cite{quine2013word}



Conversely, the intrinsic and creative value of the fact that the universality and diversity of language, as well as the possibility and impossibility of translation, maintain a balance has also been highlighted in translation theory in philosophy of language. For example, Benjamin argued that translation is not a perfect reproduction of the original text but rather an act that seeks to approach the "pure language" underlying both the original and translation, supplementing and revealing the implicit in the original~\cite{benjamin2016task}, while Derrida pointed out that the inherent différance and impossibility of exact equivalence open up space for new meanings and interpretations in translation act~\cite{derrida1985tours}. These studies suggest that the process of translation could extract new and fruitful knowledge from the fact that there are multiple languages in the world.


In the field of computational linguistics, however, the impossibility of translation is primarily regarded as an issue that requires resolution, with a paucity of studies exploring the creative dimensions of this phenomenon itself. Since the advent of computing, researchers have been developing methodologies to perform as much of the translation process as possible, which is inherently imperfect, using rule-based approaches~\cite{hutchins2007machine}, statistical methods~\cite{lopez2008statistical}, and neural network techniques~\cite{stahlberg2020neural} modeled on the human act of translation. Some studies have indeed made use of the indeterminacy of translation for specific tasks implicitly. For example, Helge Dyvik proposed that the two language models influence each other and act as a mirror~\cite{dyvik2004translations}, and studies have sought to address the scarcity of data in a few languages by complementing each other's knowledge and by aggregating the knowledge of each language to obtain information beyond that of the language with the most information~\cite{navigli2010babelnet, etzioni2007lexical}. However, their emphasis is on the variability of data across languages, and there are few examples that demonstrate a proactive approach to leveraging the intrinsic and unavoidable diversity of languages as a valuable asset, which we mentioned at the outset of this paper.

This research proposes a map to explore the huge network formed by the fact that word translations do not have a one-to-one correspondence, which is the fact that represents the impossibility of translation.
Overall, our contributions are as follows: 1. We propose \textit{Word Translation Network} directly formed from the history of human word translation. This is a network where words are nodes, and translation volume is represented as edges. 2. We propose \textit{Media of Langue}, a framework for exploring \textit{Media of Langue} for visualizing \textit{Word Translation Network}. This dictionary provides rich and intuitive linguistic and cultural knowledge that cannot be reduced to existing synonym and bilingual dictionaries while providing the functionality of existing synonym and bilingual dictionaries.
3. We release a web application as an implementation of the \textit{Media of Langue}, focusing on seven language pairs: Chinese-English, English-French, English-German, English-Japanese, English-Korean, English-Spanish, and French-Japanese. Users can access the \textit{Word Translation Network} from it.

This paper first defines a \textit{Word Translation Network} and introduces a method for constructing this network from a bilingual corpus and its unitary analysis. Next, we describe the interface so that users can explore \textit{Word Translation Network}. Finally, we describe how the proposed dictionary proposes new dictionary categories, highlighting unique features not found in existing dictionaries, and describe user experience scenarios written based on actual user usage.

\begin{figure*}[t]
\includegraphics[width=\textwidth]{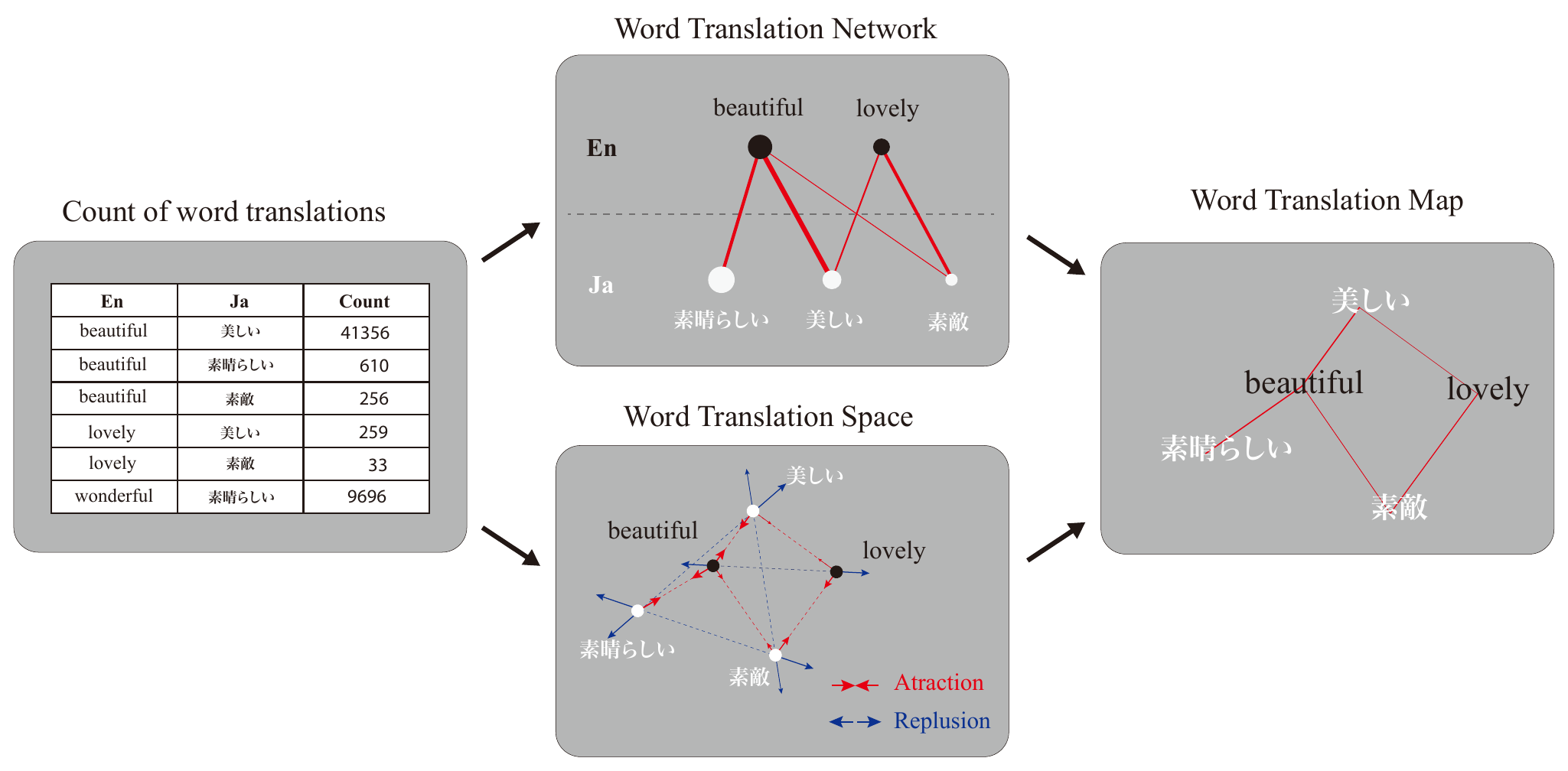}
 \caption{\textit{Word Translation Network} are formed only from information about the number of translations and then visualized in \textit{Media of Langue}}
\label{fig:forces}
\end{figure*}



\section{Related works}



\subsection{Multilingual semantic network}

In the field of semantic networks, many works have been conducted to design a network with semantic relations as edges, to bring ``meaning'' to the computer.
In the study of semantic networks, the key is where the meaning relations are extracted from and how they are defined as edges. 
Most methods extract information about semantic relationships between words from human-authored texts or databases, partition them into pre-categorized semantic relationships, and construct edges~\cite{richens1956preprogramming,chandrasekaran2021evolution,ji2021survey}.
Then, the constructed semantic networks are applied in a variety of situations, including recommendation systems~\cite{zhang2016collaborative, wang2018dkn, guo2020survey}, question answering systems~\cite{huang2019knowledge, bauer2018commonsense, song2023advancements}, and search engines~\cite{xiong2017explicit}.

Some studies have focused on developing a consistent semantic network across multiple languages using translation, whose approach has mainly taken the idea of composing monolingual semantic knowledges~\cite{navigli2010babelnet, etzioni2007lexical}.
Such a multilingual semantic network is expected to contribute strongly to the conversion, interchange, and mutual complementation of languages on the computer.
Here, we emphasize that our \textit{Word Translation Network} differs from these multilingual semantic networks in that it is not a network created by composing the semantic knowledge assumed in advance for their interchange but a network that arises directly from their interchanges themselves only and translated words are connected directly.


\subsection{Dictionary interface}

Dictionaries have been reviewed for their interfaces using the latest technology of the times, be it tablet, paper, or print.
Especially after the advent of the computer, the development of its interface was remarkable in both quality and quantity. In the early days of computers, they imitated the design of paper dictionaries, but the cross-referencing feature that existed in paper dictionaries was perfected by computers as hypertext and became more than a quantitative extension~\cite{urdang1984lexicographer,boguraev1989computational}. In a slightly later generation, on the other hand, the idea of a freer digital dictionary that took advantage of the characteristics of computers developed significantly.
\cite{klemenc2017technological} explained this development by the fact that in digital dictionaries, the database layer could be separated from the presentation layer. While the presentation layer is subject to the limitations of human cognition, the database layer does not necessarily need to be in a form that is directly recognizable to humans, but only if it is properly filtered, rearranged, and displayed in the process of transferring it to the presentation layer.
These implementation approaches are unique to computer technology, but most of them still mimic the layout of traditional paper dictionaries in their presentation layers.
Some projects, on the other hand, have not only a database structure optimized for computers but also a presentation layer with a dynamically changing map interface designed to take advantage of the strengths of digital displays~\cite{lohmann2015webvowl,mendesfollowing,khan2011data, smilkov2016embedding}. 
Their dictionary arranges words spatially by drawing the existing network/space (e.g., WorkNet~\cite{miller1995wordnet}) using computer-based network rendering technology~\cite{bannister2013force}.
Here, we emphasize that our dictionary differs from them in that it draws the new network, i.e., \textit{Word Translation Network}, in the new manner of spatial arrangement, i.e., \textit{Inter-lingual semantic space}.
This realizes a new category of dictionaries, \textit{Media of Langue} providing users with a new experience that is not found in existing dictionaries, which is described in Section~\ref{sec: features}.

\section{Word Translation Network}
\label{sec: Word Translation Network}



In this section, we introduce \textit{Word Translation Network}.
First, in Section~\ref{sec: Word Translation Network Definition}, we provide a definition of \textit{Word Translation Network}.
Next, in Section~\ref{sec: dataPreparation}, we explain the method for automatically extracting the data for \textit{Word Translation Network} construction from bilingual corpora.
Finally, in Section~\ref{sec: Quick analysis}, we present a quick analysis of \textit{Word Translation Network}.

\subsection{Definition}
\label{sec: Word Translation Network Definition}

We define the network with the following nodes and edges as \textit{Word Translation Network}.
\begin{itemize}
    \item Each node corresponds to a single word that appears in the corpora.

    The weight of the node depends on the frequency of the word in the corpora.

    \item Each edge connects two words (i.e., nodes) that appear as translation pairs in the corpora\footnote{The edges of \textit{Word Translation Network} can also be directed edges if the directionality of the translation is bound to the corpora.}.

    The weight of the edge depends on the frequency of the translation pair in the corpora.
\end{itemize}
This network can be defined for multiple language systems (i.e., all the systems in which we can believe the arbitrariness that Saussure named~\cite{de1989cours}) where an adequate amount of translations/communications between them has been accumulated.

This network forms large connected components due to the fact that one word can be translated into several words instead of just one.
We believe that all words other than proper names potentially belong to a fairly large \textit{Word Translation Network}.
As a stronger hypothesis, we also consider the possibility that all two words in the same grammatical category (except proper names) belong to the same huge network. 

Here, we emphasize that this is not a network created by the compromise of several monolingual knowledge but a network that arises directly from their interchanges between languages only.
Therefore, this ``inter-lingual'' network should be distinguished from the existing ``multi-lingual'' networks in the field of semantic networks.

Incidentally, we can extract networks representing monolingual semantic similarity relations from \textit{Word Translation Network}.
Virtual edges can be created between same-language nodes in \textit{Word Translation Network} when the two words share a common translation in another language, for example when ``beautiful'' and ``lovely'' share the Japanese translation ``utsukushi.''
Note that this is a network that changes depending on which language is paired in the generation.
In general, when semantic networks address non-categorical relations, such as nuances of semantic similarity, they encounter limitations in their expressive granularity due to the need to use categorical relations indirectly~\cite{miller1995wordnet}. On the contrary, the virtual edges in \textit{Word Translation Network} express horizontal semantic similarity based on the fact they have the same translation. 


\subsection{Data preparation}
\label{sec: dataPreparation}

\begin{figure}[t]
\centering
\includegraphics[width=0.31\textwidth]{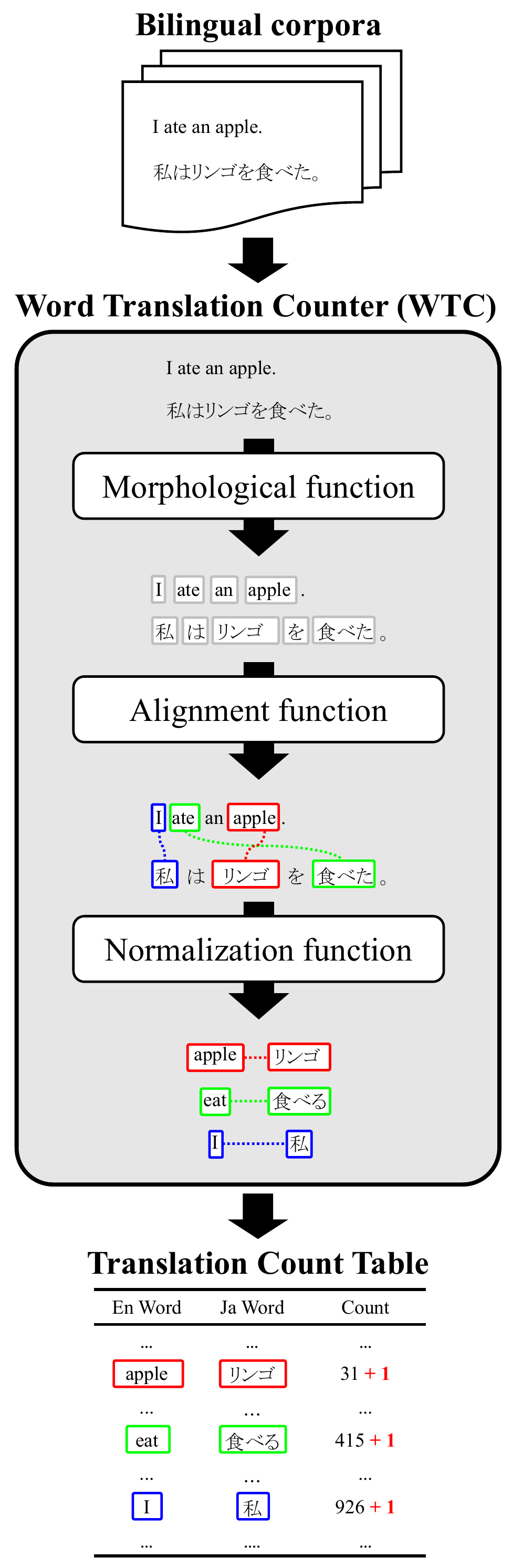}
\caption{The workflow of our WTC. Note that WTC is defined only by its inputs and outputs; its internals are implementation-dependent.}
\label{fig:LTC}
\end{figure}

\begin{table*}[t]
    \caption{Libraries used for WTC.}
    \label{table: LTC Function}
    \begin{center}
    \fontsize{9pt}{11pt}\selectfont 
    \begin{tabular*}{\textwidth}{@{\extracolsep{\fill}}lll@{}} \toprule
        Function & Library & Language / Language pair \\ \midrule
        Morphological function 
         & Jieba~\cite{jieba} & Zh \\
         & Juman++~\cite{tolmachev-etal-2018-juman} & Ja \\
         & Komoran~\cite{komoran} & Ko \\
         & NLTK~\cite{nltk} & En \\
         & SpaCy~\cite{spacy} & De, Fr, Es \\
        Alignment function 
         & AwesomeAlign~\cite{dou2021word} & De-En, En-Fr, En-Ja, En-Zh, Fr-Ja \\
         & M-BERT~\cite{devlin-etal-2019-bert} & En-Es, En-Ko \\
        Normalization function 
         & French LEFFF Lemmatizer~\cite{frenchleffflemmatizer} & Fr \\
         & Germalemma~\cite{Germalemma} & De \\
         & Juman++~\cite{tolmachev-etal-2018-juman} & Ja \\
         & Komoran~\cite{komoran} & Ko \\ 
         & NLTK~\cite{nltk} & En \\
         & SpaCy~\cite{spacy} & Es \\
         \bottomrule
    \end{tabular*}
    \end{center}
\end{table*}

\begin{table*}[t]
    \caption{Bilingual corpus used for WTC.}
    \label{table: Bilingual corpus used for LTC}
    \begin{center}
    \fontsize{9pt}{11pt}\selectfont 
    \begin{tabular*}{\textwidth}{@{\extracolsep{\fill}}ll@{}} \toprule
        Bilingual corpus & Language pair (number of used parallel sentences) \\ \midrule
        ParaCrawl~\cite{espla2019paracrawl} & De-En (20,000,000), En-Es (20,000,000), En-Fr (20,000,000), \\
        & En-Ko (4,000,000), En-Zh (20,000,000) \\
        JParaCrawl~\cite{morishita2019jparacrawl} & En-Ja (20,000,000) \\
        kittajafr~\cite{kittajafr} & Fr-Ja (6,000,000) \\
        \bottomrule
    \end{tabular*}
    \end{center}
\end{table*}

\begin{figure}[t]
    \centering
    \begin{minipage}{0.46\linewidth}
        \centering
        \includegraphics[width=\linewidth]{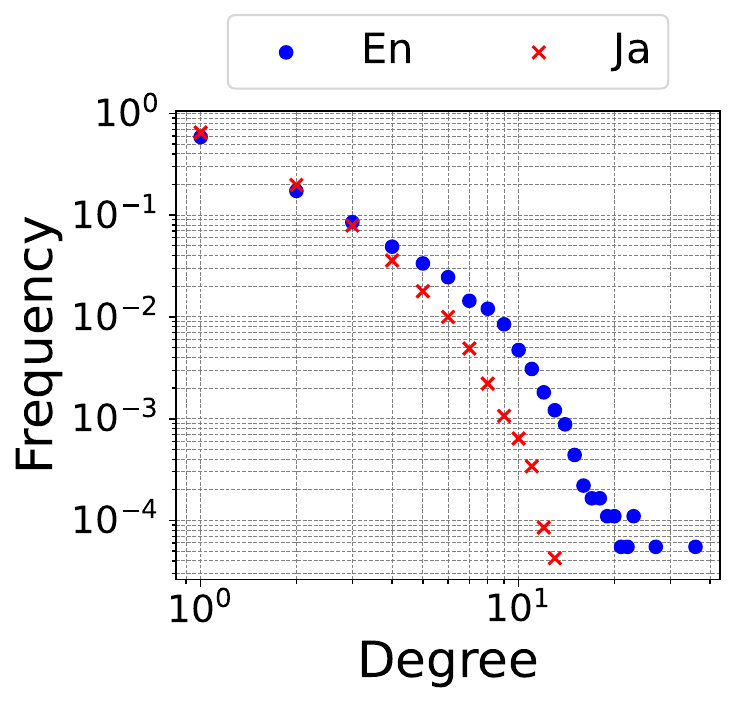}
        \subcaption{}
        \label{fig:degree_distribution_en_ja_noun_unweighted}
    \end{minipage}
    \hspace{0.01\linewidth}
    \begin{minipage}{0.48\linewidth}
        \centering
        \includegraphics[width=\linewidth]{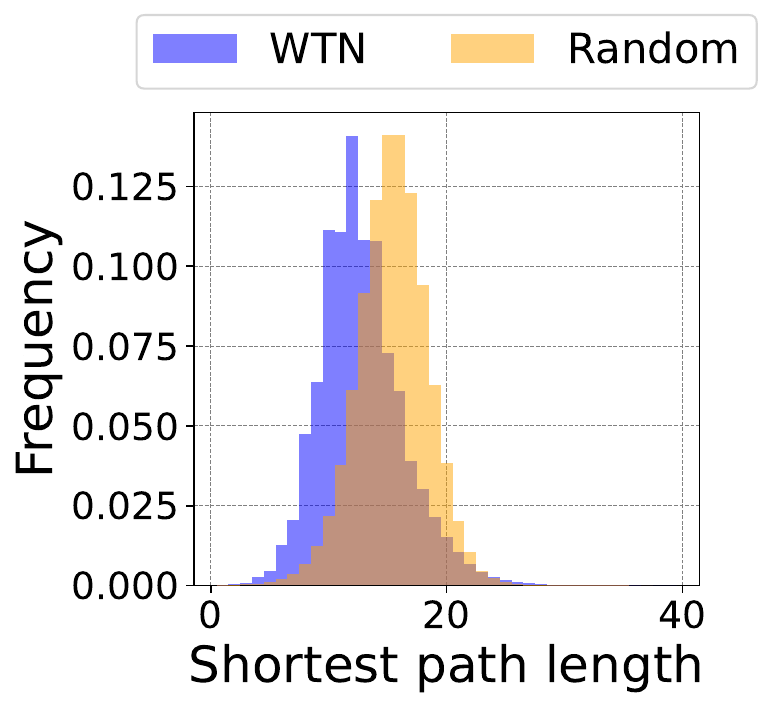}
        \subcaption{}
        \label{fig:shortest_path_length_distribution_en_ja_noun}
    \end{minipage}
    \caption{(a) Degree distribution and (b) Shortest path length distribution of \textit{Word Translation Network} with English and Japanese nouns.}
    \label{fig:combined_figures}
\end{figure}

The data needed for \textit{Word Translation Network} is the huge counts of the events ``that word was translated into this word.''
In order to obtain this data, we prepared a function, Word Translation Counter (WTC).
WTC takes bilingual corpora as an input and outputs a table that stores the number of times a word (from language 1) is translated into a word (into language 2).

WTC works by calling a morphological function, an alignment function, and a normalization function for each pair of two sentences in a translation relationship (Figure~\ref{fig:LTC}).
Currently, WTC is implemented for seven languages and seven language pairs.
The libraries used for each function are summarized in Table~\ref{table: LTC Function}.
We used ParaCrawl~\cite{espla2019paracrawl}, JParaCrawl~\cite{morishita2019jparacrawl}, and kittajafr~\cite{kittajafr} as the bilingual corpus.
The bilingual corpus and the number of used parallel sentences are summarized in Table~\ref{table: Bilingual corpus used for LTC}.
Our use of these libraries and bilingual corpora is consistent with their license and intended use.
WTC is openly available, and we expect that it will continue to be improved, including the addition of languages, by contributors from different linguistic backgrounds.

\subsection{Quick analysis}
\label{sec: Quick analysis}

Now, we present the results of a quick analysis of data obtained from WTC.
This analysis focuses on the \textit{Word Translation Network} with English and Japanese nouns (refer to the Appendix~\ref{app: detailed analysis of word translation network} for a more detailed and comprehensive analysis).

Figure~\ref{fig:degree_distribution_en_ja_noun_unweighted} shows the degree distribution of this \textit{Word Translation Network}. 
In this analysis, too infrequent translation pairs have been removed, and the network is treated as an unweighted, undirected graph. 
The points align in a straight line on the logarithmic plot, suggesting that the network exhibits scale-free properties~\cite{barabasi1999emergence}.

Figure~\ref{fig:shortest_path_length_distribution_en_ja_noun} presents the distribution of shortest path lengths in the \textit{Word Translation Network}. 
For comparison, we also include the shortest path length distribution of a bipartite random graph with the same number of nodes and edges. 
It can be observed that the shortest path lengths in the \textit{Word Translation Network} tend to be shorter than those in the random graph. 
Additionally, after calculating the bipartite clustering coefficient~\cite{latapy2008basic} for both networks, we found that the \textit{Word Translation Network} had a clustering coefficient of 0.291, whereas the random network had a slightly lower value of 0.252. 
These results show that the \textit{Word Translation Network} has a relatively short path length and high clustering characteristics, suggesting that it exhibits small-world characteristics~\cite{watts1998collective}.

\section{Media of Langue}
\label{sec: Media of Langue}

We propose \textit{Media of Langue} as the framework for exploring \textit{Word Translation Network}.
In this section, we explain the visualization of the network (Section~\ref{sec: network visualization}), then explain the design of the dictionary interface (Section~\ref{sec: User interface design}).

\subsection{Network Visualization: Word Translation Space}
\label{sec: network visualization}


In the visualization of \textit{Word Translation Network}, the nodes (i.e., words) are positioned according to the following algorithm:
\begin{enumerate}
    \item Two forces are defined:
    \begin{itemize}
        \item Repulsion between words in the same language system.
        \item Attraction between translated words in different language systems.
    \end{itemize}
    \item The positions of words are determined by minimizing the energy associated with these forces.
\end{enumerate}
Although this method is similar to previous work on network visualization~\cite{bannister2013force}, the two forces can be explained by linguistic intuition as follows: Repulsion corresponds to the force of a word to secure its value by its difference from other words, or the fact that ``words (or translations for a word) with the same meaning are not needed,'' and attraction corresponds to the force of a translation to suppress differences from the target word, or the fact that ``translation should preserve meaning.'' 
The magnitude of each force depends on the frequency of occurrence in the corpora (a detailed formulation, including the normalization method, is provided in the Appendix). 

Here, we emphasize that this is not a space created by the compromise of several monolingual knowledge but a space that arises directly from their interchanges between languages only.
Therefore, this ``inter-lingual'' space should be distinguished from the existing ``multi-lingual'' spaces in the field of semantic spaces and could be termed \textit{Word Translation Space}, or \textit{Inter-Lingual Semantic Space}.
However, it is a future task to analyze the characteristics of this space as a semantic space through comparison with existing monolingual/multilingual semantic spaces.


Based on the coordinates of this space, the nodes of the network are drawn as the text of the corresponding word, and the edges as lines.
The word size and line width are determined by inheritance from the node weight and the edge weight, respectively.





\subsection{User interface design}
\label{sec: User interface design}
The UI of \textit{Media of Langue} consists of two parallel layers: the \textit{Word Translation Network} layer, which visualizes the \textit{Word Translation Network}, and the material corpora layer, which displays the bilingual parallel corpus in response to the other layer. 
Note that these two layers can be seen as extensions of the concept of ``langue'' and ``parole''~\cite{de1989cours}.
The explanations in this section are based on the actual \textit{Media of Langue} we have implemented as a web page.

\paragraph{\textit{Word Translation Network} layer}

This layer (the left side of Figure~\ref{fig:mediaOfLangue}) visualizes \textit{Word Translation Network} based on the method described in Section~\ref{sec: network visualization}.
\textit{Word Translation Network} consisting of several hundred words centered on the words that the user is focusing on is visualized, helping the user to recognize the network.
Users can choose to view the network in either 2D or 3D. 
In 2D, they can quickly get an overview of the network at a glance, while in 3D, they can more accurately perceive the semantic distances and network structure between words by rotating the network.

\paragraph{Material Corpus layer}
The parallel corpora that served as material for the \textit{Word Translation Network} layer are displayed (the right side of Figure~\ref{fig:mediaOfLangue}). 
In a pair of vertically aligned sentences of the corpora, the words in the focused translation relationship are emphasized by being connected with lines, just as they were in \textit{Word Translation Network}.
When users click on a word or translation line in the \textit{Word Translation Network} layer, the corresponding sentences from the corpora immediately flow into this layer.
These corpora serve as the so-called example sentences of the dictionary. 
This feature ensures the transparency of this dictionary, as it means that the sole basis of the dictionary is always displayed in the interface.


\paragraph{Exploration}
There were two types of search methods. The first was a word search. When a word of interest was entered in the search window, a map centered on the word expanded. The users could select language pairs, part of speech, dimension, and initial language. The second was a search that moved sequentially by interacting with the map. When the user clicks on a word, the map moves so that the word is in the center. The map can also be zoomed in and out.
These two methods allow users to explore the vast landform that exists at the boundary between languages in the same way that anyone can explore virtually anywhere on earth from a personal display. 


\section{Feature Analysis}
\label{sec: features}

\textit{Media of Langue} is the first interface to explore the huge network formed by past translation practices based on translation indeterminacy. Here, however, we explain how the ability to explore this network inevitably gives rise to properties that other dictionaries do not have, based on their framework and implementation.

\subsection{Structural features}

\textit{Media of Langue} has the following structural features based on the described implementation.

\paragraph{Consolidation and extension of existing multiple dictionaries} \textit{Media of Langue} encompasses multiple types of dictionaries and simultaneously provides information equivalent to multiple entries in each dictionary. For example, when a map is created using two languages, it serves as both the two synonym dictionaries for each language and the bilingual dictionary between the two languages; thus, it plays the role of three dictionaries at the same time. 
In addition, the content covered by multiple entries in the three dictionaries is now contained on a single screen so that the information can be viewed at a glance without having to flip through pages. This not only saves time and effort but also brings richer knowledge that can only be gained from a global perspective.


\paragraph{Coordinate space embedded with familiar linguistic senses} In \textit{Media of Langue}, words in one language can be captured relative to their network/spatial distances from multiple words in the other language. This allows users to capture the nuances of the foreign language using the linguistic senses of their native language. In addition, users who are proficient in multiple languages can capture the nuances of an unfamiliar language while using multiple familiar languages.

\paragraph{Direct reflection of the actual state of translation} 
The paths connecting words on the map can be interpreted as a ``translation traffic network'' that records translation events, and the width of the path can be interpreted as a ``translation traffic volume''. This map gives the user a realistic picture of the translation traffic, not only which translations are considered appropriate, but also how words in the two languages are actually being translated into each other around the world.

\paragraph{Uncovering implicit and global semantic continuities}
The \textit{Media of Langue} does not use explicit knowledge, but reflects the translator's nonverbal understanding of the meaning behind the translation activities.
This allows the dictionary to expose a continuum of meanings based on latent human universals beyond languages. It can suggest words that are out of scope at the usual synonym but actually have a continuity of meaning as synonyms in the broadest sense, or it can tell us how one concept is connected to another.


\subsection{Illustrative examples for user experience}

Here we describe the user experience scenario written based on actual user usage. This scenario is a longer experience (1-3 minutes) designed to give an overview of many of the possible experiences of \textit{Media of Langue}, and in fact, any part of this scenario can be omitted. This scenario was created by the authors by asking 50 users of the app how they use it, reconstructing them. 

Assume a user who is an English speaker and a learner of Japanese. 

Interested in the difference between ``正直'' and ``誠実,'' both of which translations of ``honest,'' she goes to \textit{Media of Langue}, and types ``honest'' in the search window.
The map containing words such as ``honest,'' ``sincere,'' ``truthful,'' ``frank,'' ``straightforward,'' ``正直,'' ``誠実,'' ``素直,'' ``率直,'' ``真摯'' then unfolds.
At this point, the basic functions of a normal bilingual dictionary are fulfilled by looking up words directly linked with lines to ``honest,'' such as ``正直,'' ``誠実,'' ``素直,'' ``率直.'' The dictionary also fulfills the basic functions of an English synonym dictionary for ``honest'' by focusing only on English, and a Japanese synonym dictionary for ``正直/誠実,'' too.

Looking at the map, she first notices that ``正直'' is quite close to ``honest'' (the translation is a relatively one-to-one correspondence), whereas there is no English word that corresponds exactly to ``誠実.'' Seeing that ``誠実'' is located near the center of the triangle of ``honest,'' ``sincere,'' and ``truthful'' (slightly closer to ``honest''), she imagines the nuance of ``誠実'' as a new blended nuance from the nuances of three words that have already acquired a rich linguistic sense in the appropriate proportions.
Then, to check the actual usage of the word, she clicks on the thickest line near the center, between ``honest'' and ``誠実'', then actual sentences in which ``honest'' and ``誠実'' are translated appear on the right side of the screen.
She then clicks on other lines to check the usage, confirming that the nuances she understands spatially match the usage, and adding a dependency on the situation to her knowledge, such as the fact that ``正直'' tends to be used in more casual situations than ``誠実.''

Furthermore, looking down, she notices that in the direction of ``frank,'' words containing the nuance of simplicity such as ``frank,'' ``素直,'' and ``straightforward'' are spreading, while toward ``sincere,'' words with nuances related to truth such as ``truthful,'' ``true,'' and ``genuine,'' are spreading.
When she clicks on the word ``素直,'' a space appears in which words related to obedience, such as ``compliant,'' ``obedient,'' and ``docile,'' float.
It reminds her that ``honesty,'' which is often required and valued, is next to ``obedience,'' which is often betrayed, deceived, and criticized.

In addition to the cases examined in detail above, many other situations can be considered. 

\begin{itemize}
\item A user who has only thought of the word ``implication'' as a synonym for ``meaning'' will find through \textit{Media of Langue} not only the usual synonym dictionary words such as ``signification'' and ``context'' but also the words that are seemingly distant concepts such as ``sense,'' ``importance,'' and ``consciousness,'' via the Japanese words. 
\item A user familiar with English and Japanese can acquire the nuance of ``beau'' in French, a third language, based not only on the relationship with ``beautiful,'' ``cute,'' and ``lovely'' in English but also on the relationship with ``美しい,'' ``綺麗,'' and ``素敵'' in Japanese.
\end{itemize} 



\section{Conclusion}

This paper presented \textit{Media of Langue}, a framework for exploring \textit{Word Translation Network}.
\textit{Word Translation Network} is a huge network where words are nodes, and translation volume is represented as edges, which was formed by past language translations on the balance between language universality and diversity.

We defined this network, explained the method of constructing it from a parallel corpus, and performed a preliminary analysis of the network we had created.
We then explained that we can form a new kind of dictionary that refers to the semantic place of many words at once with a chain of mutual translation by creating a map that explores this network. This map serves as a panoramic and nuanced bilingual and synonym dictionary but also suggests our unique engagement with language and meaning, which is not reducible to the traditional use of dictionaries.

``Media of Langue'' is a ``medium'' between \textit{langue}  and \textit{langue} but, therefore,  a ``medium'' between humans and \textit{langue}. 
We hope that this dictionary will be used across the world and that it will become an open platform that continues to reflect the possibilities/impossibilities of communication, and the universality/diversity of humanity.

\section{Limitations}
Since this paper focuses on the proposal of a new dictionary category, \textit{Media of Langue}, a detailed analysis of \textit{Word Translation Network}, which we also propose, has not been carried out. In particular, it is important to evaluate, under specific tasks, how the novelty of focusing only on the translation history between languages, without using inter-lingual information, quantitatively differs from the existing multilingual semantic network/space. Also, regarding the evaluation of the interface, this paper focused on the structurally generated features of this new dictionary category, but a qualitative analysis of the actual user experience with this application is a future task.
There are still technical limitations with regard to the stage of obtaining the history of word translations, i.e., ``that word was translated into this word,'' which depend on the insufficient amount of parallel corpora and the accuracy of the alignment libraries. However, future contributions of the NLP community to corpus collection and alignment methods will greatly improve the number of language-pair and the quality of the map can be improved. 
We hope that the map will eventually become huge one that reflects many of the myriad translations that exist around the world.
\section{Potential risks}
The example sentences in \textit{Media of Langue} displayed in the corpus layer are the corpus data itself, which is the raw material for the map, unlike most existing dictionaries that use example sentences selected by experts. As such, they are not necessarily representative of a given translation and may contain slang or direct expressions. However, the security gained by filtering these out is a trade-off for how well the dictionary reflects actual language traffic. The latter is also an important feature of this category of dictionaries. Careful decisions about this policy will need to be made in future distributions of this dictionary.
On the other hand, while the project confronts the fact that there exist multiple languages and re-proposes its value, it relies on the quantity and quality of the translations that are actually done, which leads to a bias towards languages with high speaker and recognition levels. In fact, all of the seven languages currently deployed are English on one side, with the exception of Japanese-French. It is possible to form a map between languages for which there is no actual translation data by adding English, but this is not \textit{Media of Langue} based on actual direct translations between the two languages, but an indirect one, heavily influenced by English and its culture, which is problem with the English pivot in general. For the dictionary to fully realise its multilingualist nature, there needs to be a conscious effort to actively focus on the languages of minority speakers, or more precisely, between the languages of minority traffic.

\bibliography{references}

\clearpage
\appendix

\section{Formulate \textit{Inter-lingual semantic space}}

Here we formulate the \textit{Inter-lingual semantic space}.
Let $\mathcal{L}$ be the set of languages, $\mathcal{W}^{(l)}_\mathrm{all}$ be the set of words of the language $l$, and $\mathbf{x}_{w}$ be the coordinates of the word $w$.
The word coordinates are optimized to minimize the loss $E_\mathrm{total} = E_\mathrm{rep} + E_\mathrm{att}$, where $E_\mathrm{rep}$ is the energy loss due to repulsive forces between words of the same language, and $E_\mathrm{att}$ is the energy loss due to attractive forces between words of different languages;
\begin{align}
\label{equ:rep loss}
    E_\mathrm{rep} 
    &= \sum_{l \in \mathcal{L}} 
    \sum_{\substack{
        u,v \in \mathcal{W}^{(l)}_\mathrm{all} \\
        u \neq v
    }}
    \mathrm{rep}(\mathbf{x}_{u}, \mathbf{x}_{v}; u, v), \\
    E_\mathrm{att}
    &= \sum_{\substack{
        l_1, l_2 \in \mathcal{L} \\
        l_1 \neq l_2
    }}
    \sum_{\substack{
        u \in \mathcal{W}^{(l_1)}_\mathrm{all} \\
        v \in \mathcal{W}^{(l_2)}_\mathrm{all}
    }}
    \mathrm{att}(\mathbf{x}_{u}, \mathbf{x}_{v}; u, v).
\end{align}
$\mathrm{rep}$ and $\mathrm{att}$ are functions that return the energy loss between the two words due to repulsion and attraction, respectively.
We used the metaphor of the elastic force of a spring and the repulsive force of an electric charge, referring to previous work in network visualization~\cite{bannister2013force}:
\begin{align}
\label{equ:rep func}
    \mathrm{rep}(\mathbf{x}_{u}, \mathbf{x}_{v}; u, v) &= 
    \frac{
        q_u q_v
    }{
        \| \mathbf{x}_{u} - \mathbf{x}_{v} \|_{2}
    }, \\
    \mathrm{att}(\mathbf{x}_{u}, \mathbf{x}_{v}; u, v) &= 
    k_{u, v} \| \mathbf{x}_{u} - \mathbf{x}_{v} \|_{2}^{2},
\end{align}
where $q_u$ and $k_{u, v}$ are the ``charge'' of the word $u$ and the ``spring constant'' of the spring connecting the word $u$ and $v$, respectively.
Here, note that repulsive forces, which are usually defined for all nodes, only work between nodes of the same language in our method. 

\section{Exploration of network}

Here, we describe the two algorithms we have developed to enable users to explore \textit{Word Translation Network} composed of the ``neighborhood words'' of the user-searched word.

\subsection{Computing with a subset of words}
\label{sec: algorithm using a subset of words}

We have to address the following three important factors to provide users with well-organized \textit{Word Translation Network} of ``neighborhood words'': 
(i) the corpora size varies widely among different language pairs, 
(ii) the frequency of word occurrences varies widely among different words, 
(iii) the words particularly relevant to the user-searched word should be highlighted.
Here, let $\mathcal{W}^{(l)}$ be the ``neighborhood words'' in language $l$, and $\mathcal{P}$ be the user-specified set of language pairs.

First, to deal with (i), we normalize the number of occurrences of the word $u$, $c_u$, and the number of occurrences of the translation between $u$ and $v$, $c_{u, v}$, as follows:
\begin{align}
    \bar{c}_u &= \frac{1}{\left| \mathcal{L}_\mathrm{pair}^{(l_u)}  \right|} 
            \sum\nolimits_{l_v \in \mathcal{L}_\mathrm{pair}^{(l_u)} }
            \sum\nolimits_{v \in \mathcal{W}^{(l_v)}_\mathrm{all} }
            \frac{ c_{u, v} }{ T_{l_u, l_v} }, \\
    \bar{c}_{u, v} &= \frac{ c_{u, v} }{ T_{l_u, l_v} },
\end{align}
where
\begin{align}
    \mathcal{L}_\mathrm{pair}^{(l_u)} &= \{ l ~ | ~ (l_u, l) \in \mathcal{P} \}, \\
    T_{l_u, l_v} &= 
        \sum\nolimits_{s \in \mathcal{W}^{(l_u)}_\mathrm{all} }
        \sum\nolimits_{t \in \mathcal{W}^{(l_v)}_\mathrm{all} }
        c_{s, t}.
\end{align}
$T_{l_u, l_v}$ is a divisor for handling variations in corpora size across different language pairs.

Next, to address (ii), we obtain the ``original'' charge and spring coefficients, $\bar{q}_u$ and $\bar{k}_{u,v}$, with the following normalization:
\begin{equation}
    \bar{q}_u = \frac{ f \left(
            \bar{c}_u
        \right)
    }{C}, ~~
    \bar{k}_{u,v} = \frac{ f \left(
            \bar{c}_{u, v}
        \right)
    }{C},
\end{equation}
where
\begin{align}
    C &= \frac{
        \sum\limits_{ (l_1, l_2) \in \mathcal{P} } ~ 
        \sum\limits_{
            \substack{
                s \in \mathcal{W}^{(l_1)}\\
                t \in \mathcal{W}^{(l_2)}
            }
        }
        ~ p(s;\mathbf{x}_s) p(t;\mathbf{x}_t)
        f \left(
            \bar{c}_{s, t}
        \right)
    }{
        \sum\limits_{ (l_1, l_2) \in \mathcal{P} } ~ 
        \sum\limits_{
            \substack{
                s \in \mathcal{W}^{(l_1)}\\
                t \in \mathcal{W}^{(l_2)}
            }
        }
        ~ p(s;\mathbf{x}_s) p(t;\mathbf{x}_t)
    }, \\
    f(c) &= c^{\gamma}.
\end{align}
The function $p$ weights the ``original'' charge and spring coefficients according to the relevance to the user-searched word (we define $p$ in the next paragraph).
$C$ is a divisor to maintain an appropriate scale, and $f$ is a function to mitigate the influence of words or pairs that appear excessively frequently.
$\gamma$ is a constant between 0 and 1 that governs the nonlinearity ($\gamma = 0.5$ in our implementation).

Finally, to address (iii), we define the actual charge and spring coefficients, $q_u$ and $k_{u,v}$, by weighting the ``original'' charge and spring coefficients:
\begin{align}
    q_u &= \bar{q}_u \cdot p(u;\mathbf{x}_u), \\
    k_{u,v} &= \bar{k}_{u,v} \cdot p(u;\mathbf{x}_u) p(v;\mathbf{x}_v),
\end{align}
where
\begin{equation}
\label{equ:function p}
    p(u;\mathbf{x}_u)=(1 - \alpha_{t} ^ {t_u}) \cdot \alpha_{x} ^ {\| \mathbf{x}_u \|_{2}}.
\end{equation}
$t_u$ is the time elapsed since the word $u$ is added to $\mathcal{W}$, and $\alpha_{t}$ and $\alpha_{x}$ are positive hyperparameters that adjust the temporal and spatial weighting, respectively ($\alpha_{t}=0.9985$ and $\alpha_{x}=0.5$ in our implementation).
This gives more weight to words that have been displayed for a longer time and that are closer to the user-searched word.

\subsection{Selecting words for coordinate calculation}
\label{sec: selecting words for coordinate calculation}

$\mathcal{W}$ is first a set consisting of only one word searched by the user, and then words are successively added to and deleted from this set, following the assumption that ``the tentatively computed word coordinates correctly represent (to some extent) the semantic distribution of the words around the central word.''
Based on this assumption, words connected to words close to the searched word are added, and words that have moved away from the searched word are deleted.

The word additions and deletions are performed when the word coordinates have approximately converged.
Word addition is performed by adding a word that is connected to ``tentatively near-centered words'', i.e., ``parent'' words.
Word deletion is performed according to two rules: (i) words that are not ``parent'' words and are not connected to ``parent'' words are deleted, (ii) if the number of words exceeds $N_\mathrm{max}$ (150 in our implementation), the words farthest from the center are deleted.

\section{Detailed Analysis of Word Translation Network}
\label{app: detailed analysis of word translation network}

\begin{table*}[p]
    \caption{The proportion of nodes in the largest connected component. The values in parentheses indicate the number of nodes in the largest component and the total number of nodes.}
    \label{table:max_size_of_connected_component}
    \begin{center}
    \begin{tabular}{@{}lllll@{}} \toprule
     & Adjective & Adverb & Noun & Verb \\ \midrule
    De-En & 0.592 (4,460/7,538) & 0.855 (627/733) & 0.569 (17,744/31,159) & 0.877 (5,872/6,696) \\ 
    En-Es & 0.272 (3,021/11,103) & 0.639 (679/1,063) & 0.543 (14,458/26,631) & 0.748 (7,093/9,480) \\ 
    En-Fr & 0.416 (4,218/10,133) & 0.713 (917/1,286) & 0.522 (16,769/32,109) & 0.773 (5,843/7,561) \\ 
    En-Ja & 0.664 (3,262/4,909) & 0.895 (649/725) & 0.475 (19,812/41,704) & 0.867 (7,643/8,820) \\ 
    En-Ko & 0.651 (164/252) & 0.767 (394/514) & 0.521 (8,229/15,784) & 0.786 (1,137/1,446) \\ 
    En-Zh & 0.622 (2,499/4,020) & 0.844 (146/173) & 0.685 (9,435/13,776) & 0.901 (4,508/5,006) \\ 
    Fr-Ja & 0.671 (1,778/2,649) & 0.970 (460/474) & 0.614 (11,702/19,067) & 0.872 (4,177/4,788) \\ 
    \bottomrule
    \end{tabular}
    \end{center}
\end{table*}

\begin{figure*}[p]
\centering
\includegraphics[width=0.78\textwidth]{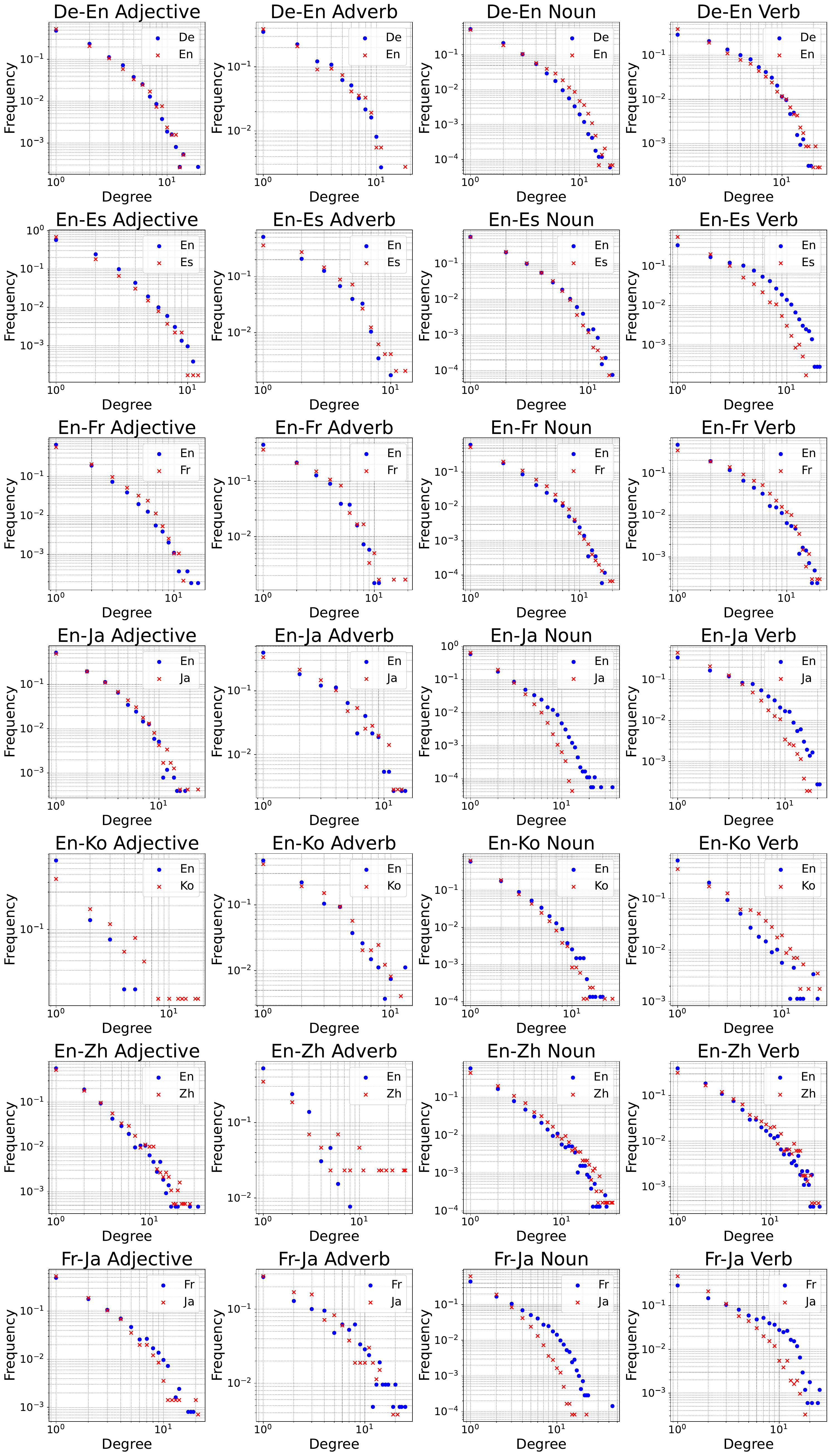}
\caption{Degree distribution of each \textit{Word Translation Network}.}
\label{fig:degree_distributions}
\end{figure*}

\begin{table*}[p]
    \caption{The bipartite clustering coefficient~\cite{latapy2008basic} in \textit{Word Translation Network} (approximated). The values in parentheses indicate those in a random network with the same number of nodes and edges.}
    \label{table:approximated_bipartite_clustering_coefficient_with_random_graph}
    \begin{center}
    \begin{tabular}{@{}lllll@{}} \toprule
     & Adjective & Adverb & Noun & Verb \\ \midrule
    De-En & 0.287 (0.226 ± 0.012) & 0.284 (0.187 ± 0.012) & 0.265 (0.236 ± 0.020) & 0.254 (0.173 ± 0.009) \\ 
    En-Es & 0.329 (0.255 ± 0.021) & 0.299 (0.218 ± 0.017) & 0.322 (0.233 ± 0.015) & 0.403 (0.213 ± 0.008) \\ 
    En-Fr & 0.302 (0.246 ± 0.023) & 0.294 (0.210 ± 0.008) & 0.247 (0.254 ± 0.021) & 0.211 (0.197 ± 0.013) \\ 
    En-Ja & 0.303 (0.239 ± 0.018) & 0.262 (0.190 ± 0.015) & 0.291 (0.252 ± 0.017) & 0.264 (0.195 ± 0.014) \\ 
    En-Ko & 0.410 (0.280 ± 0.031) & 0.290 (0.218 ± 0.016) & 0.237 (0.242 ± 0.017) & 0.288 (0.201 ± 0.009) \\ 
    En-Zh & 0.258 (0.222 ± 0.011) & 0.317 (0.235 ± 0.015) & 0.247 (0.207 ± 0.014) & 0.210 (0.141 ± 0.005) \\ 
    Fr-Ja & 0.328 (0.220 ± 0.010) & 0.237 (0.132 ± 0.003) & 0.348 (0.236 ± 0.019) & 0.284 (0.188 ± 0.009) \\ 
    \bottomrule
    \end{tabular}
    \end{center}
\end{table*}

\begin{table*}[p]
    \caption{The average shortest path length in the largest connected component of each \textit{Word Translation Network} (approximated). The values in parentheses indicate those in a random network with the same number of nodes and edges.}
    \label{table:approximated_average_shortest_path_length_with_random_graph}
    \begin{center}
    \begin{tabular}{@{}lllll@{}} \toprule
     & Adjective & Adverb & Noun & Verb \\ \midrule
    De-En & 11.67 (11.05 ± 0.16) & 6.92 (6.25 ± 0.17) & 11.22 (13.12 ± 0.12) & 7.70 (7.64 ± 0.07) \\ 
    En-Es & 13.13 (15.70 ± 0.30) & 8.62 (8.01 ± 0.21) & 12.54 (14.28 ± 0.15) & 9.42 (9.24 ± 0.15) \\ 
    En-Fr & 12.61 (13.92 ± 0.45) & 9.04 (7.69 ± 0.20) & 11.35 (14.49 ± 0.30) & 8.50 (8.61 ± 0.09) \\ 
    En-Ja & 9.51 (9.92 ± 0.20) & 6.89 (6.12 ± 0.12) & 12.79 (15.66 ± 0.20) & 8.76 (8.44 ± 0.09) \\ 
    En-Ko & 5.51 (6.50 ± 0.45) & 6.78 (6.64 ± 0.20) & 10.64 (13.86 ± 0.24) & 6.54 (6.97 ± 0.13) \\ 
    En-Zh & 7.78 (9.15 ± 0.12) & 4.05 (4.36 ± 0.21) & 7.94 (9.63 ± 0.11) & 6.30 (6.39 ± 0.04) \\ 
    Fr-Ja & 8.91 (8.77 ± 0.14) & 5.05 (4.39 ± 0.05) & 11.62 (11.24 ± 0.10) & 7.28 (7.11 ± 0.07) \\ 
    \bottomrule
    \end{tabular}
    \end{center}
\end{table*}

\begin{figure*}[p]
\centering
\includegraphics[width=0.78\textwidth]{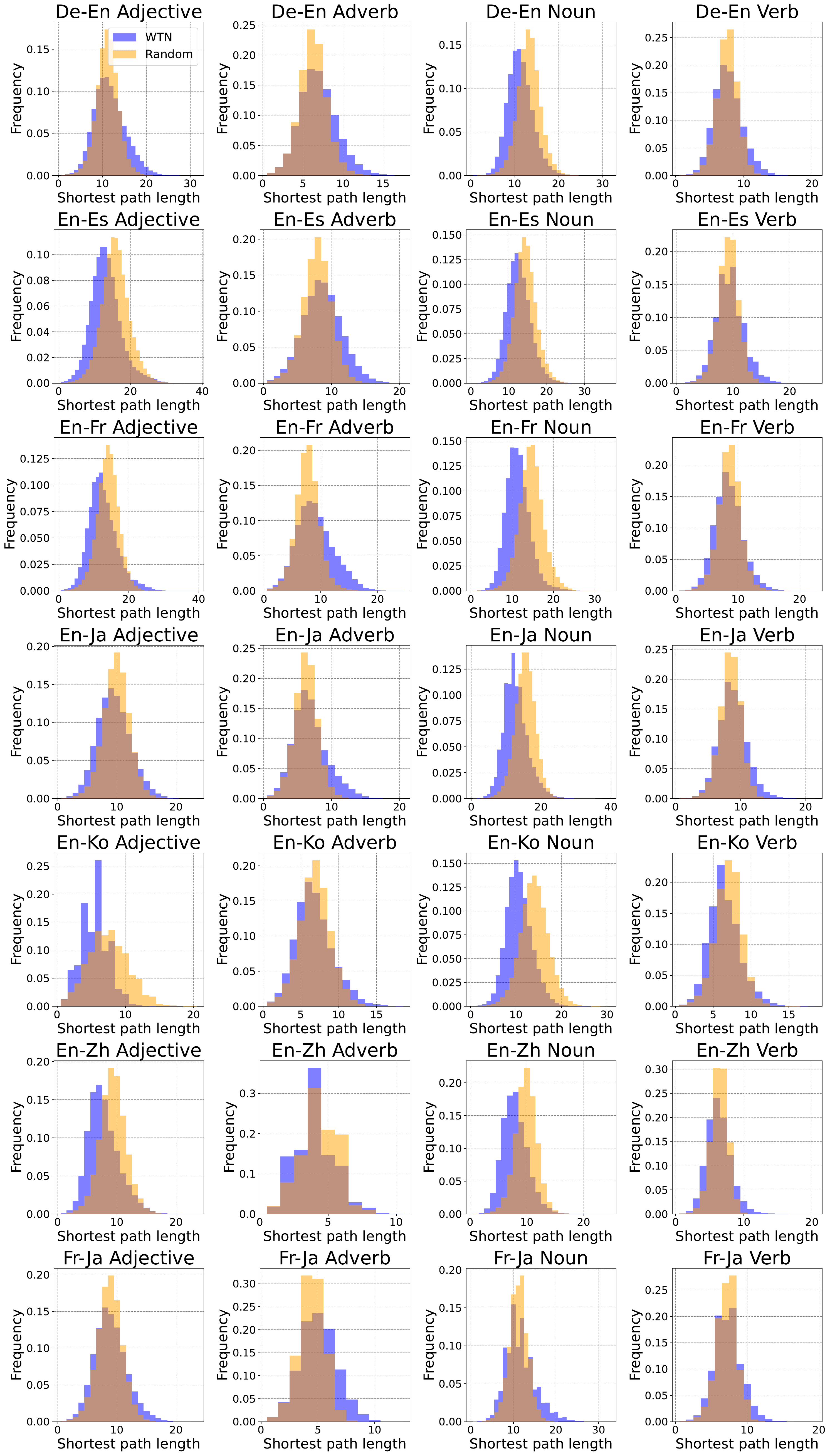}
\caption{Distribution of shortest path lengths in the largest connected components of each \textit{Word Translation Network} (approximated).}
\label{fig:shortest_path_length_distribution}
\end{figure*}

Here, we provide a more detailed and comprehensive analysis of the \textit{Word Translation Network}. 
All analysis was conducted on a Linux machine equipped with an Intel\textsuperscript{\textregistered}~Core\texttrademark{}~i9-11900H CPU @ 2.50\,GHz and 62\,GB of memory.
Our analysis delves into the network structure and key properties, offering insights that were not fully covered in the main text.

As a preliminary, we offer some assumptions. 
In this analysis, translation pairs that occur too infrequently have been removed, and the network is modeled as an unweighted, undirected graph. 
Specifically, edges between nodes $u$, whose number of occurrences in the corpus is $c_u$, and $v$, whose number of occurrences in the corpus is $c_v$, are omitted if $c_{u,v}$, the number of occurrence of their translation in the corpus, satisfies the inequality $c_{u,v} \leq \alpha \min\left(c_{u}, c_{v}\right) + \beta$, where $\alpha = 0.008$ and $\beta = 5$.
This filtering process ensures that the analysis focuses on meaningful and frequent word translations.

Additionally, we define the construction method for the bipartite random network used as a comparison throughout this section. This bipartite random network contains the same number of nodes and edges as the \textit{Word Translation Network}. Let the number of nodes in language 1 and language 2 be $N_1$ and $N_2$, respectively, and the number of edges be $M$.
The random bipartite network is generated by selecting $M$ edges uniformly at random from the $N_1N_2$ possible edges between the two sets of nodes. This random network serves as a baseline for comparison in terms of degree distribution, clustering coefficient, and shortest path lengths.

First, Table~\ref{table:max_size_of_connected_component} displays the proportion of nodes that belong to the largest connected component in each \textit{Word Translation Network}. The values in parentheses show the actual number of nodes in the largest component and the total number of nodes in the network. From this table, it is evident that for most language pairs and parts of speech, more than half of the words are part of the same connected network, indicating a high degree of connectivity.

Next, Figure~\ref{fig:degree_distributions} presents the degree distribution of each \textit{Word Translation Network}. The data points tend to align linearly on a logarithmic plot, suggesting that the degree distribution follows a power-law, which is characteristic of scale-free networks. This observation implies that a few nodes act as hubs, connecting a large number of other nodes, while most nodes have relatively few connections.

Table~\ref{table:approximated_bipartite_clustering_coefficient_with_random_graph} provides the bipartite clustering coefficients for each \textit{Word Translation Network}. Since the conventional clustering coefficient would yield a value of zero for bipartite graphs, we use a clustering coefficient specifically designed for bipartite networks~\cite{latapy2008basic}. For computational efficiency, we approximate this value by averaging the clustering coefficients of a random sample of 100 nodes. The table also shows the average clustering coefficient for 10 randomly generated bipartite networks with the same number of nodes and edges. Notably, the clustering coefficient of the \textit{Word Translation Network} is consistently higher than that of the random bipartite networks. This suggests that words in the \textit{Word Translation Network} tend to form tightly connected clusters, a hallmark of small-world networks.

Table~\ref{table:approximated_average_shortest_path_length_with_random_graph} summarizes the average shortest path length in the largest connected component of each \textit{Word Translation Network}, with corresponding values for random bipartite networks. Figure~\ref{fig:shortest_path_length_distribution} shows the distribution of shortest path lengths. The shortest path lengths of the \textit{Word Translation Networks} tend to be as short as, or shorter than, those of the random networks, further supporting the idea that these networks exhibit small-world properties~\cite{watts1998collective}. This indicates that, on average, words are more closely related to each other in translation networks than would be expected by chance.

In conclusion, the detailed analysis of the \textit{Word Translation Network} reveals several key structural properties: the network is scale-free, exhibits small-world characteristics, and forms clusters more densely than random bipartite graphs. These findings provide valuable insights into the organization and efficiency of word translations between languages.




\end{document}